\documentclass[letterpaper, 10 pt, conference]{ieeeconf}  

\usepackage[T1]{fontenc}

\IEEEoverridecommandlockouts                              

\usepackage{graphics} 
\usepackage{epsfig} 
\usepackage{mathptmx} 
\usepackage{times} 
\usepackage{amsmath} 
\usepackage{amssymb}  
\usepackage{multirow}

\usepackage{hyperref}
\usepackage{tablefootnote}
\usepackage{threeparttable}

\title{\LARGE \bf
Failure Forecasting Boosts Robustness \\ of Sim2Real Rhythmic Insertion Policies
}

\author{Yuhan Liu$^{1}$, Xinyu Zhang, Haonan Chang, and Abdeslam Boularias$^{2}$
\thanks{The authors are with the Department of Computer Science, Rutgers
University, 08854 New Jersey, USA. Email: }
\thanks{$^{1}$Yuhan Liu, {\tt\small yl1834@rutgers.edu}}%
\thanks{$^{2}$Abdeslam Boularias, {\tt\small ab1544@cs.rutgers.edu}}%
}

\begin{document}
\newcommand{\ours}{POFR}
\newcommand{\OURS}{\textbf{POFR} }

\maketitle
\thispagestyle{empty}
\pagestyle{empty}

\begin{abstract}

This paper addresses the challenges of Rhythmic Insertion Tasks (RIT), where a robot must repeatedly perform high-precision insertions, such as screwing a nut into a bolt with a wrench. The inherent difficulty of RIT lies in achieving millimeter-level accuracy and maintaining consistent performance over multiple repetitions, particularly when factors like nut rotation and friction introduce additional complexity. We propose a sim-to-real framework that integrates a reinforcement learning-based insertion policy with a failure forecasting module. By representing the wrench’s pose in the nut’s coordinate frame rather than the robot’s frame, our approach significantly enhances sim-to-real transferability. The insertion policy, trained in simulation, leverages real-time 6D pose tracking to execute precise alignment, insertion, and rotation maneuvers. Simultaneously, a neural network predicts potential execution failures, triggering a simple recovery mechanism that lifts the wrench and retries the insertion. Extensive experiments in both simulated and real-world environments demonstrate that our method not only achieves a high one-time success rate but also robustly maintains performance over long-horizon repetitive tasks. For more information please refer to the website: \href{https://jaysparrow.github.io/rit}{jaysparrow.github.io/rit}.
\end{abstract}


\section{Introduction}

\begin{figure}[tp]
    \centering
    \includegraphics[width=0.9\linewidth]{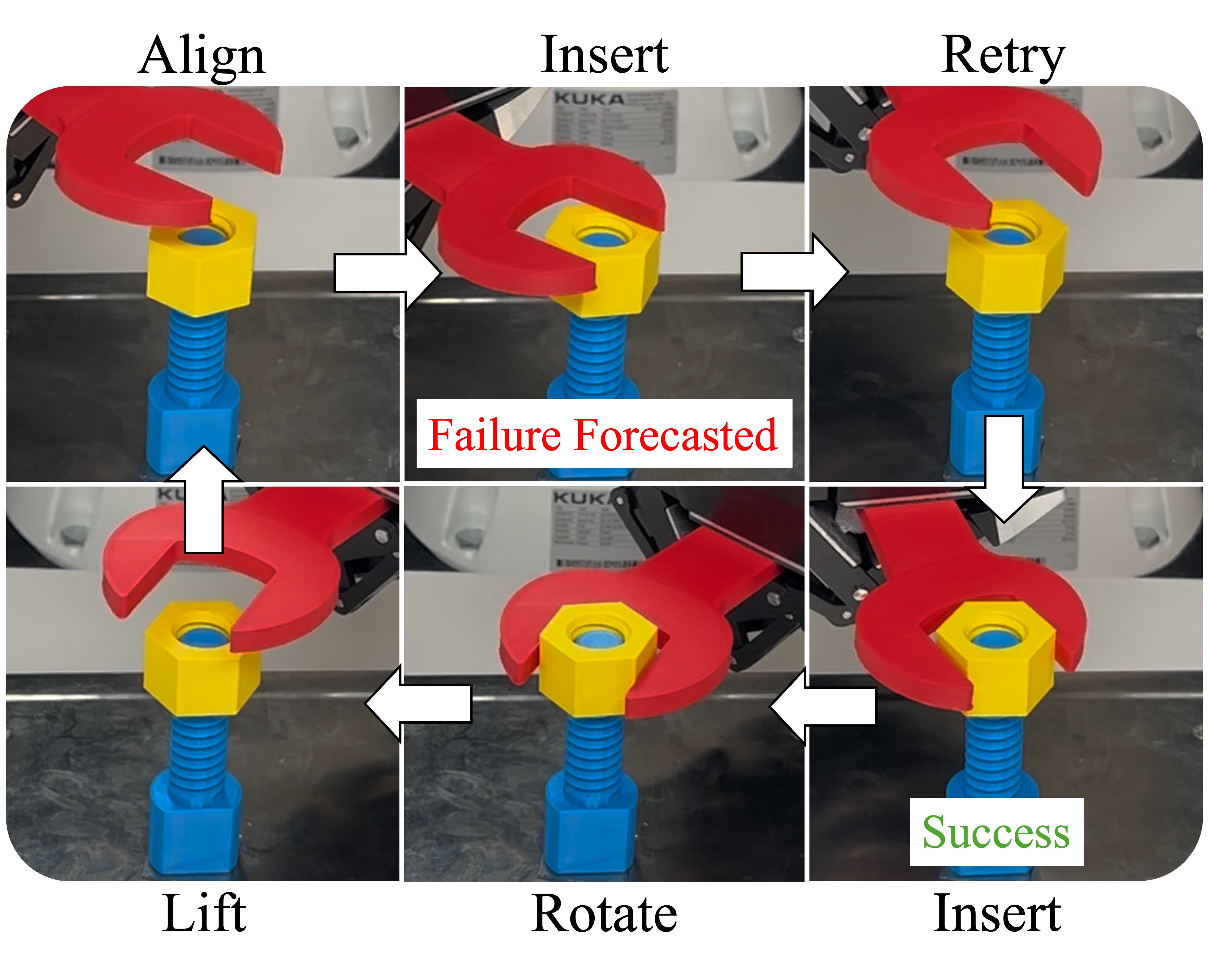}
    \caption{Robot executing wrench-based nut insertion: employing a trained neural network to forecast potential failures, the system autonomously corrects its trajectory and retries until the nut is fully inserted, exemplifying enhanced robustness of rhythmic insertion policies.}
    \label{fig:overview}
\end{figure}

A Rhythmic Insertion Task (\textit{RIT}) is a type of manipulation task that requires performing precise insertion actions multiple times in succession. It has a wide range of applications, from industrial assembly to everyday tasks. In \textit{RIT}, a robot must repeatedly insert an object into a receptacle within a single task. A typical example is using a wrench to screw a nut into a bolt, a common industrial assembly task. As shown in Fig.~\ref{fig:overview}, the robot must repeatedly execute three key steps: aligning the wrench with the nut, pushing down (inserting), and rotating. In this paper, we focus on solving \textit{RIT} by studying the task of screwing a nut into a bolt using a wrench to better understand the challenges and solutions for general repeated insertion tasks.

There are two major challenges in \textit{RIT}. The first challenge is achieving high-precision insertion. Insertion tasks require millimeter accuracy. Unlike traditional insertion tasks, where the receptacle is fixed, in tasks such as screwing a nut into a bolt, the nut is not fixed but can rotate along the bolt due to friction. This added degree of freedom significantly increases the difficulty of insertion. The second challenge is ensuring consistent success over multiple repetitions. A single successful insertion is not enough—long-horizon performance requires a robust failure recovery mechanism. Without proper failure recovery, even if the one-time success rate exceeds 90\%, the overall success rate collapses rapidly as the number of repetitions increases.

To address \textit{RIT} effectively, a robust single-insertion policy and a reliable recovery strategy are essential. Research on one-time insertion tasks has identified three main approaches: (1) visual-tracking alignment with predefined trajectory searching~\cite{chang2024insert, jiang2022state}, (2) visual-tracking alignment combined with learning-based searching~\cite{tang2023industrealtransferringcontactrichassembly, beltranhernandez2022acceleratingrobotlearningcontactrich, zhang2021learninginsertionprimitivesdiscretecontinuous}, and (3) end-to-end visual policies~\cite{nair2023learning, spector2021insertionnet, lee2020making}. While the tracking-based predefined trajectory method is transferable, it is limited by fixed search patterns, and end-to-end visual policies require extensive fine-tuning on real-world data. A hybrid approach using learning-based strategies with a 6D pose tracker balances adaptability and sim-to-real transferability. Based on this insight, we build on top of \textit{IndustReal}~\cite{tang2023industrealtransferringcontactrichassembly} for the challenging task of screwing a nut in a bolt with a wrench. We train a pose-based insertion policy in simulation, where nut and wrench poses are directly available, and deploy it in the real world using the 6D pose tracker \textit{FoundationPose}~\cite{wen2024foundationposeunified6dpose}. Moreover, we find that representing the wrench's pose in the nut's coordinate frame—rather than the robot's frame—substantially improves sim-to-real performance by a large margin.

However, achieving a high one-time success rate is not sufficient for \textit{RIT}. Due to the repetitive nature of the task, a failure forecasting and recovery strategy is crucial. Failure recovery mechanisms have been widely studied in long-horizon manipulation~\cite{brunke2022safe}. Failure recovery consists of two key components: the recovery policy and the recovery triggering-events. Some approaches explicitly collect failure data to train a recovery policy~\cite{reichlin2022back, vats2024recoverychaining}. However, for \textit{RIT}, a simple recovery strategy—lifting and retrying the insertion—is often sufficient. In this paper, we adopt this straightforward recovery policy. Regarding failure detection, there are two common approaches: anomaly detection~\cite{thananjeyan2021recovery, srinivasan2020learning} and failure prediction~\cite{luo2024sample, farid2022failurepredictionstatisticalguarantees}. Anomaly detection models normal system behavior and identifies deviations, making it suitable when ample success data is available. In contrast, failure prediction, which directly predicts if a trajectory is going to end in a failure, is more appropriate for our case since we have both success and failure data from simulation training. In practice, we train a network to detect the potential failure within the current execution, and perform the lift-up recovery once a failure is predicted.

In summary, we propose a sim-to-real policy in object coordinates with failure recovery. The proposed method has two components: (1) a reinforcement learning policy for single-insertion tasks that takes the wrench’s pose in the nut’s coordinate frame as input and outputs the 6D displacement of the wrench, and (2) a failure recovery module that lifts the wrench to retry the insertion when needed. In real-world experiments, we use the 6D pose tracker \textit{FoundationPose}~\cite{wen2024foundationposeunified6dpose} to track object and wrench poses and a failure forecasting method based on simulation data to predict potential failures. Extensive experiments validate the effectiveness of our method in both simulation and real environments.

Our contributions can be summarized as follows: 
\begin{enumerate}
    \item To the best of our knowledge, the proposed method is the first robust policy specifically designed to address the repeated insertion task.
    \item We improve the sim-to-real transferability by modifying \textit{IndustReal}’s coordinate representation from the robot frame to the receptacle’s frame. Extensive experiments in both simulation and real-world environments validate the effectiveness of this modification
    \item We systematically evaluate multiple failure forecasting methods and their impact on \textit{RIT} performance through extensive experiments in both simulation and real-world settings.
\end{enumerate}

\section{Related Works}
\textbf{Tight Insertion}. Object insertion under (sub-)millimeter tolerances has been extensively studied in the literature. Data-driven approaches learn end-to-end insertion policies from high-dimensional visual inputs~\cite{nair2023learning, spector2021insertionnet, lee2020making}. These methods usually require extra online fine-tuning~\cite{nair2023learning} or other sensor modalities~\cite{spector2021insertionnet, lee2020making} to deploy in the real-world. Tracking-based methods often demonstrate stronger transfer ability in real-world deployment. Combinations of various searching patterns and force-feedback controllers for insertion have also been studied~\cite{chang2024insert, jiang2022state}. While fixed patterns are resilient to calibration errors and external disturbances on stationary receptacles, they are not robust over moving objects (e.g., rotating nuts) as in our task. Recently, RL has became a popular choice for insertion policy learning~\cite{tang2023industrealtransferringcontactrichassembly, beltranhernandez2022acceleratingrobotlearningcontactrich, zhang2021learninginsertionprimitivesdiscretecontinuous}. Among these works, IndustReal~\cite{tang2023industrealtransferringcontactrichassembly} exhibits good performances by introducing a dense signed-distance-function-based reward and a distance-based curriculum. Our non-recovery insertion policy is built by modifying IndustReal's robot-frame observation to the object's frame. This simple change, in addition to failure forecasting and recovery, leads to a significant performance gain.

\textbf{Failure Recovery}. Failures frequently happen in robotics applications, and a recovery behavior is often necessary for improving the system robustness~\cite{brunke2022safe}. A common approach is to learn a recovery policy for taking the system back to the nominal state distribution from offline demonstration data~\cite{reichlin2022back} or through online interaction with an environment~\cite{vats2024recoverychaining}. The behavior of these learning-based recovery policies is unpredictable due to the black-box nature of the deep neural networks, which may bring an extra complexity to the system. Instead, we adopt a simple yet very effective  recovery strategy - autonomous resetting and trying again. Another series of works~\cite{liu2022robot, liu2023interactive, liu2024model, ren2023robotsaskhelpuncertainty} rely on humans intervention, and the human feedback collected during deployment can be further used for improving the original policy~\cite{liu2023interactive}. In contrast, our method is completely automatic, and thus more appropriate for contact-rich manipulation tasks that require frequent recovery. 

\textbf{Failure Prediction} Many prior efforts have focused on failure detection, where a potential execution failure is predicted by a different module from the actor policy. The literature on failure detection mainly falls into two domains: out-of-distribution (OOD) detection and failure classification. OOD detection aims to identify inputs or events that deviate from the nominal inputs. Within robotics applications, different ways of measuring the nominal distribution have been proposed: \cite{thananjeyan2021recovery, srinivasan2020learning} approximate a constraint function from data while \cite{lee2020guided, mitsioni2021safe, wu2021rlad, lutjens2019safe} estimate the region of safety or uncertainty for collision avoidance. In the case of the state-based high-precision assembly tasks considered in this work, the effective workspace is usually confined within a small area, and a significant outlier state (if any) should appear much later because action errors accumulate slowly. Due to the time-sensitive nature of our task, OOD detection can result in retrying at such a late time that the remaining duration is short for any further actions. The other series of works intend to explicitly foresee a future failure for an early intervention. Traditional methods utilize control barrier functions~\cite{ames2016control} or Hamilton-Jacobi reachability analysis~\cite{akametalu2014reachability} to check if the state at the next step falls in a failure set before actually executing an action. Despite the performance guarantees offered by these methods, they are not applicable to our task as an explicit uncertainty definition is required, which is unavailable in the real-world deployment. Recently, techniques from conformal prediction~\cite{luo2024sample} and probably approximately correct (PAC) learning~\cite{farid2022failurepredictionstatisticalguarantees} have been employed for error-bounded failure prediction. However, these methods assume the same distribution of the deployment environment as the training data, which is not the case in our sim-to-real application because the physical characteristics (e.g. friction and object deformation) of the real world greatly deviate from the simulated environments. In contrast, our failure prediction models assume only the same deployment policy without relying on environmental assumptions.

\section{Problem Formulation}
We consider each single round of \textit{RIT}, where a robot is tasked to insert a grasped tool (e.g. a wrench) into a predefined pose relative to a target object (e.g. a nut). The goal pose of the tool relative to the object is given as ${P^*}_{tool}^{obj}\in \mathbb{SE}(3)$. At every time-step $t\in [0, T]$, the robot estimates from RGB-D data the state $s_t=[P_{tool}^{cam}(t), P_{obj}^{cam}(t)]$ which is the pose of the tool and the object in the camera frame. The estimated pose is denoted as observation $o_t=[\hat{P}_{tool}^{cam}(t), \hat{P}_{obj}^{cam}(t)]$. The objective is to learn a policy $\pi(o_t)$ that outputs the transformations of the tool's pose, denoted as action $a_t\in\mathbb{SE}(3)$, so that it reaches the predefined goal at some time $T_S\in [0, T)$,
\begin{equation}
    P_{tool}^{cam}(T_S) = {P^*}_{tool}^{obj} P_{obj}^{cam}(T_S).
\end{equation}
Notice that this tool-object alignment task is more challenging than the regular insertion task. First, it is harder to maintain the tool at the aligned pose than in a regular insertion task. Second, both the tool and target objects move when they are in contact with each other (e.g. the nut will rotate around the bolt when the wrench moves on its surface). Third, the errors accumulate in a repeated tool-use application (such as removing a nut with repeated wrench actions) and a slight drop of the alignment success rate can lead to a large decline in the overall task success rate.

\section{Method}
\begin{figure}
    \centering
    \includegraphics[width=0.8\linewidth]{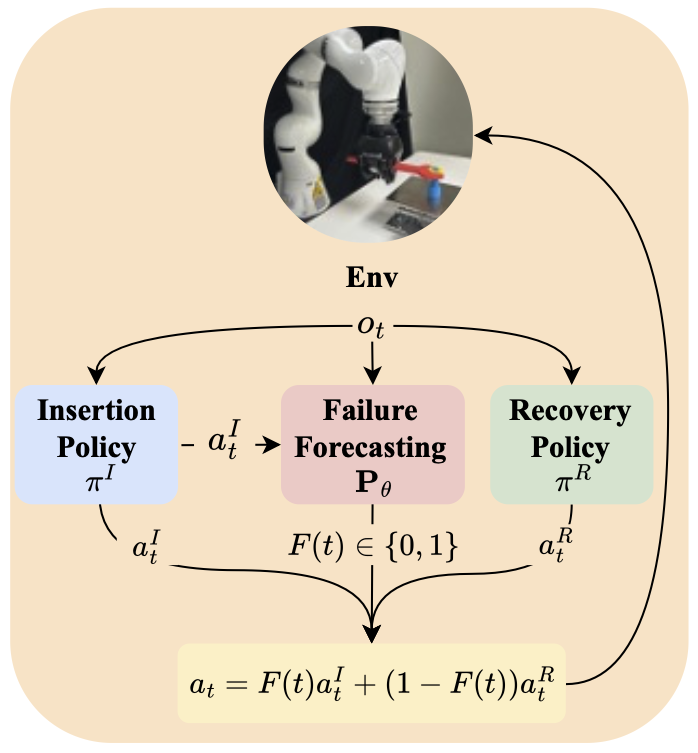}
    \caption{System Pipeline}
    \label{fig:system_pipeline}
\end{figure}
We present a robust closed-loop system for tool use, with a focus on the task of using a wrench to insert/remove a nut into a bolt.
The proposed system, depicted in Fig.~\ref{fig:system_pipeline}, contains a module for predicting failures and recovering by trying the tool-target alignment again. The following sections introduce each module. Sec. \ref{method:alignment} summarizes the non-recovery RL policy for insertion. Sec. \ref{method:recovery} describes our simple, yet effective, strategy for recovering from out-of-distribution states. Sec. \ref{method:failure} details three different models for failure forecasting under a unified probabilistic framework. Lastly,  Sec. \ref{method:sim2real} describes the deployment of the system trained in simulation to a real-world setup. 

\subsection{Insertion Policy}\label{method:alignment}
The insertion policy is trained with the Proximal Policy Optimization (PPO)~\cite{schulman2017proximalpolicyoptimizationalgorithms} algorithm in simulated environments in Isaac Gym. We adopt the signed-distance field rewards and sampling-based curricula from IndustReal~\cite{tang2023industrealtransferringcontactrichassembly}, but we modify the observation and action representations for better generalization and easier sim-to-real deployment.

Specifically, the observations in IndustReal are the current and goal poses of the end-effector in the world frame, as well as the joint positions of the robotic arm. An action $a_t \in \mathbb{SE}(3)$ is a displacement of the end-effector at a given time-step. Instead, we adopt an \textbf{object-centric} representation, where an observation  $o_t$ is the current and goal poses of the tool relative to the object, $o_t=[{\hat P}_{tool}^{obj}(t),{P^*}_{tool}^{obj}]$, and an action $a_t$ is the displacement of the tool in the object's frame $a^{I}_t=\Delta {P}_{tool}^{obj}(t)$. This object-centric representation is embodiment-agnostic and initialization-invariant. Neither the initial pose of the object in the world nor the initial pose of the tool in the gripper matter, as long as the pose of the tool relative to the object is within the training distribution. As shown in our experiments, this simple change improves the policy's performance by an astonishingly large margin. This performance gain could be attributed to  the smaller observation space in our representation, compared to IndustReal's, which results in a better data coverage given the same training time.

\subsection{Recovery Strategy} \label{method:recovery}
The learned insertion policy follows a general search pattern. It often fails to produce an accurate insertion when an out-of-distribution state is encountered due to bad initialization or unexpected physical interactions, and also due to the compounding errors over time. By analyzing the behavior of the policy in the real-world setup, we found that a failure happens in three cases: (1) the tool (e.g., wrench) is stuck on the object due to friction, (2) the object (e.g., nut) rotates when the tool is moving and the policy cannot recover from the rotation misalignment, (3) the initial pose is out of distribution. By simply moving the tool away from the object and re-starting, all of these failure cases can be addressed. 

We define the actions of the recovery policy as follows. At each time-step $t$, the recovery policy $\pi^R$ plans a path for the tool object from its current pose towards the the predefined pre-insertion pose (refer to "Retry" in Fig. \ref{fig:overview}). The path is interpolated by sub-goals and the $\mathbb{SE}(3)$ distance between consecutive  sub-goals is bounded by the same action step-size of the insertion policy $\pi^I$. The robot then moves the tool to the first sub-goal and completes this step. The recovery policy, once being triggered, continuously runs for $T_R$ steps, and then hands over the control back to the insertion policy.

\subsection{Failure Forecasting}\label{method:failure}
\begin{figure}
    \centering
    \includegraphics[width=\linewidth]{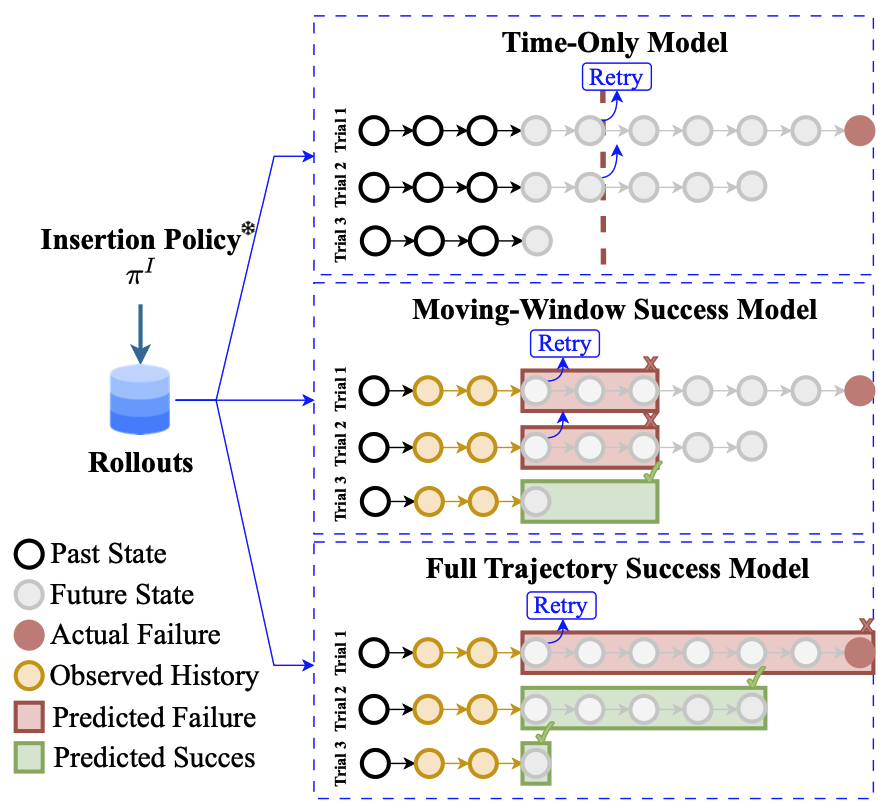}
    \caption{Proposed Failure Forecasting Models}
    \label{fig:failure_forecasting}
\end{figure}
Given a trained insertion policy $\pi^I$, the failure forecasting problem can be tackled in two steps: (1) learning the probability of successfully aligning the tool within a given time window, conditioned on a history of observed features, and (2) monitoring this probability over time and triggering the recovery policy whenever the success probability falls below a given threshold.

Formally, let us define the random variable {\it success} as the event of successfully inserting the tool onto the object, and let
$T_S\in [0, \infty)$ denote the time when this happens. Starting at time-step $t$, we define the probability of success occurring in the next $T_F$ time-steps as:
\begin{equation}
    p_t \stackrel{def}{=} \mathbf{P}_\theta(T_S \in [t, T_F] | f_t, \pi^I),
    \label{main_model}
\end{equation}
where $f_t$ is a feature vector of the history until time $t$. The trajectory-level failure forecasting model $\mathbf{F}: t\rightarrow \{0, 1\}$ predicts at time $t$ a future failure  when the success probability is lower than a predefined threshold $\alpha \in [0, 1]$. In other terms, 
$F(t)=1 $ if $p_t<\alpha$ and $F(t)=0 $ if $p_t\geq\alpha$.

In the remaining of this section, we present three variants of the probabilistic model $\mathbf{P}_\theta$, as shown in Fig.~\ref{fig:failure_forecasting}. The same policy $\pi^I$ is used in the three variants. 

\textbf{Time-Only Model} The simplest model is the empirical distribution of the success time of the trajectories sampled from the policy:
\begin{equation}
    p_t = \mathbf{P}_\theta(T_S \in [t, T] | t, \pi^I) \approx \frac{\# (T_S \in [t, T])}{|\mathcal{T}_t|},
\end{equation}
where $\mathcal{T}_t$ denotes the set of training sub-trajectories starting at time-step $t$, collected with policy $\pi$, and $\# (T_S \in [t, T])$ indicates the number of such sub-trajectories that succeeded within $T$ time-steps. Note that this is a special case of the model in Eq.~\ref{main_model} with $T_F = T$ and $f_t = t$. This model assumes that the success probability is independent of the observation history. In practice, we found that setting the  cut-off threshold $\alpha$ to $0.2$ provides a good balance between insertion attempts and recovery behavior.

\textbf{Moving-Window Success Model}\label{Method:forecasting2} The three failure cases described in Sec.~\ref{method:recovery} reveal that it is possible to forecast failures based on the history of observed tool and object poses in a given trial. Specifically, we train a neural network with the dataset of trajectories collected using policy $\pi^I$ to fit the probability distribution of a success event happening in a time window in the future $[t,T_F]$, where $T_F$ is predefined. 
\begin{eqnarray}
    p_t &=& \mathbf{P}_\theta(T_S \in [t, T_F] | o_{t-T_H:t}, t, \pi^I) \\
     &=& S(t| o_{t-T_H:t}, t, \pi^I) - S(t+T_F| o_{t-T_H:t}, t, \pi^I),
\end{eqnarray}
Here, $o_{t-T_H:t}$ is a history of received observations, with predefined length $T_H$, and $S(\tau| o_{t-T_H:t}, t, \pi^I)$ is the probability that a success happens any time in the future after time-step $\tau$, where $\tau\geq t$. 

In the survival analysis literature, $S$ is called the {\it  survival function} and is given as a {\it Weibull} distribution. In this work, we define the feature vector $f_t$ at time-step $t$  as $[o_{t-T_H:t}, t]$ and train a three-layer Multi-Layer Perceptron (MLP) that takes as input the feature vector $f_t$ and returns the scale and shape parameter $\theta_t = \big(\lambda(f_{t}), \rho(f_{t})\big)$ of the Weibull distribution at time-step $t$. The survival probability is then given as: 
\begin{equation}
\begin{aligned}
S(\tau| f_t, \pi^I) = \exp\big(-(\frac{\tau}{\lambda(f_t)})^{\rho(f_t)}\big).
\end{aligned}
\end{equation}
The neural network that returns parameters $\lambda$ and $\rho$ as functions of input features $f_t$ is trained offline by maximizing the log likelihood under the  Weibull distribution of the success times in the training data collected by rolling the insertion policy $\pi^I$.
The training data contains $10,000$ episodes collected in simulation. The samples are labeled by their success time (which is set to the maximum time $T$ if failure occurred), and a binary indicator marking all failure episodes as censored. In our wrench-nut system, most of the success events happen between $50$ and $100$ steps. 

\textbf{Full-Trajectory Success Model} The previous model aims to predict the success time. Instead, we consider predicting a simpler target here; the success of the trajectory before the maximum termination time $T$:
\begin{equation}
\begin{aligned}
p_t = \mathbf{P}_\theta(T_S \in [t, T] | o_{t-T_H:t}, t, \pi^I)
\end{aligned}
\end{equation}
If the failure threshold $\alpha$ equals $\frac{1}{2}$, this model degenerates to a binary classification task. For training, each sample of the dataset described above is re-labeled with a binary success indicator, and the data labeled as failures are up-sampled to match the number of successes. A two-layer MLP is trained to classify trajectories as success or failure given input features $f_t= [o_{t-T_H:t}, t]$. 

\subsection{Sim-to-Real Transfer}\label{method:sim2real}
The object-centric insertion policy was trained on IsaacGym for the Kuka iiwa 14 manipulator equipped with a Robotiq 3-finger gripper, using the Task-Space Impedance Controller. The policy is zero-shot deployed  in the real-world with the same emobodiment but a different position controller that uses force feedback to avoid strong contact forces that can damage the setup. The seamless sim-to-real transfer is due to the proposed object-centric representation. 

An RGB-D-based pose tracker, FoundationPose~\cite{wen2024foundationposeunified6dpose}, is utilized to obtain the poses of the tool and the object in real-world. To accommodate for the tracking noises and the control inaccuracies, randomization and noises are applied to the poses during training in simulation. The initial position of the tool relative to the object is randomized within a range of $\pm1 cm$ in the x-y plane and $[0.5, 1]cm$ in the z-axis. The relative yaw rotation is sampled from a range of $\pm10^\circ$. The observed relative poses are added with xyz-position noises and yaw-rotation noise sampled uniformly from $\pm2mm$ and $\pm10^\circ$, respectively. These randomization and noises make the task in simulation more difficult than in the real world.

\section{Experiments}
\begin{table*}
\centering  
\begin{threeparttable}
    \caption{Insertion Success Rates and Total Number of Time Steps in Simulation  Without Wrench-Nut Friction}
    \begin{tabular}{lcccccccccc}
    \hline
    
    \multirow{2}{*}{Method} & \multicolumn{2}{c}{Size-5} & \multicolumn{2}{c}{Size-4} & \multicolumn{2}{c}{Size-3} & \multicolumn{2}{c}{Size-2} & \multicolumn{2}{c}{Size-1} \\
    \cline{2-3}\cline{4-5}\cline{6-7}\cline{8-9}\cline{10-11}
     & Succ. Rate & Steps & Succ. Rate & Steps & Succ. Rate & Steps & Succ. Rate & Steps & Succ. Rate & Steps \\
     
    \hline 
    IndustReal~\cite{tang2023industrealtransferringcontactrichassembly} & $82.8\pm1.7\%$ & $106$ 
    & $71.9\pm5.2\%$ & $110$ 
    & $59.8\pm1.4\%$ & $119$ 
    & $72.3\pm3.4\%$ & $95$ 
    & $39.1\pm4.7\%$ & $94$ \\

    Object-Centric\tnote{1} & $93.2\pm0.6\%$ & $80$ 
    & $98.0\pm0.7\%$ & $73$ 
    & $97.9\pm1.2\%$ & $65$ 
    & $97.9\pm1.2\%$ & $65$ 
    & $94.1\pm2.0\%$ & $68$ \\

    \hline
    Time-Only\tnote{2} & $92.6\pm2.0\%$ & $85$ 
    & $98.6\pm0.6\%$ & $75$ 
    & $97.9\pm0.6\%$ & $66$ 
    & $98.0\pm1.2\%$ & $65$ 
    & $94.1\pm2.1\%$ & $68$ \\

    Moving-Window\tnote{2} & $\mathbf{97.9\pm1.5\%}$ & $84$ 
    & $\mathbf{99.2\pm1.0\%}$ & $76$ 
    & $\mathbf{98.4\pm1.0\%}$ & $68$ 
    & $98.8\pm0.7\%$ & $68$ 
    & $\mathbf{95.9\pm1.8\%}$ & $75$ \\

    Full-Trajectory\tnote{2} & $97.5\pm1.2\%$ & $91$ 
    & $97.1\pm0.3\%$ & $80$ 
    & $98.2\pm1.2\%$ & $72$ 
    & $\mathbf{99.0\pm0.3\%}$ & $70$ 
    & $95.1\pm2.6\%$ & $85$ \\
    
    \hline
    \end{tabular} 
    \label{tab:low-friction}
    \begin{tablenotes}
       \item [1] IndustReal with an object-centric representation described in Sec.~\ref{method:alignment}. No recovery mechanism.
       \item [2] Insertion policy with a recovery mechanism guided by the corresponding failure forecasting model in Sec.~\ref{method:failure}, built over Object-Centric.
     \end{tablenotes}
\end{threeparttable}
\end{table*}

\begin{table*}
    \centering    
    \caption{Insertion Success Rates and Total Number of Time Steps  in Simulation With Wrench-Nut Frictions}
    \begin{tabular}{lcccccccccc}
    \hline
    
    \multirow{2}{*}{Method} & \multicolumn{2}{c}{Size-5} & \multicolumn{2}{c}{Size-4} & \multicolumn{2}{c}{Size-3} & \multicolumn{2}{c}{Size-2} & \multicolumn{2}{c}{Size-1} \\
    \cline{2-3}\cline{4-5}\cline{6-7}\cline{8-9}\cline{10-11}
     & Succ. Rate & Steps & Succ. Rate & Steps & Succ. Rate & Steps & Succ. Rate & Steps & Succ. Rate & Steps \\
     
    \hline 
    IndustReal~\cite{tang2023industrealtransferringcontactrichassembly} & $73.6\pm3.0\%$ & $114$ 
    & $63.3\pm3.4\%$ & $117$ 
    & $48.2\pm5.3\%$ & $109$ 
    & $59.6\pm5.0\%$ & $105$ 
    & $32.0\pm4.4\%$ & $96$  \\

    Object-Centric & $88.3\pm3.3\%$ & $89$ 
    & $91.6\pm1.5\%$ & $80$ 
    & $89.8\pm3.9\%$ & $72$ 
    & $95.5\pm2.0\%$ & $72$ 
    & $89.1\pm0.6\%$ & $80$  \\

    \hline
    Time-Only & $90.4\pm2.2\%$ & $98$ 
    & $96.5\pm0.9\%$ & $91$ 
    & $95.5\pm1.4\%$ & $81$ 
    & $96.7\pm1.2\%$ & $75$ 
    & $90.8\pm1.5\%$ & $87$  \\

    Moving-Window & $93.0\pm1.2\%$ & $99$ 
    & $\mathbf{96.9\pm1.1\%}$ & $89$ 
    & $96.1\pm2.3\%$ & $81$ 
    & $97.5\pm1.8\%$ & $76$ 
    & $91.4\pm2.0\%$ & $85$  \\

    Full-Trajectory & $\mathbf{94.3\pm1.8\%}$ & $101$ 
    & $96.5\pm0.9\%$ & $90$ 
    & $\mathbf{96.9\pm1.1\%}$ & $81$ 
    & $\mathbf{98.0\pm0.9\%}$ & $78$ 
    & $\mathbf{92.4\pm1.8\%}$ & $98$  \\
    
    \hline
    \end{tabular} 
    \label{tab:high-friction}
\end{table*}

Our empirical study aims to answer two questions: (1) can the object-centric representation improve the generalization ability of the insertion policy and provide zero-shot sim-to-real transfer? and (2) can the recovery strategy guided by the failure forecasting neural network boost the system's robustness for the repeated high-precision task? We conduct extensive experiments on the single-time wrench-nut alignment task in both simulation and real-world, and demonstrate the significance of the improved system's robustness on the highly challenging nut insertion/removal task.

\subsection{Setup}\label{exp:setup}
\begin{figure}
    \centering
    \includegraphics[width=0.8\linewidth]{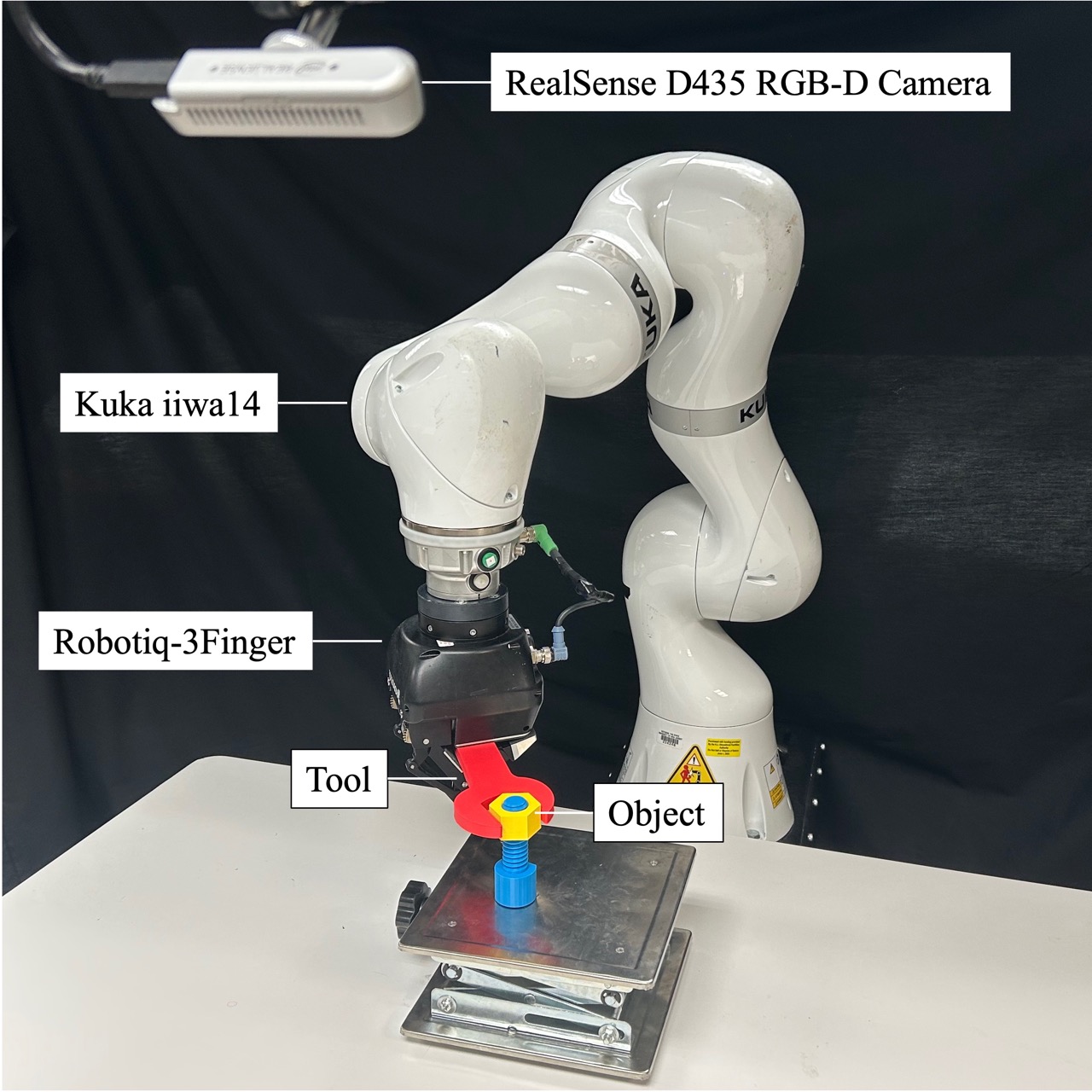}
    \caption{Robot setup: A Kuka iiwa14 manipulator equipped with a Robotiq-3Finger gripper. A RealSense D435 RGB-D camera was used for visual tracking.}
    \label{fig:real-setup}
\end{figure}
\begin{figure}
    \centering
    \includegraphics[width=0.5\linewidth]{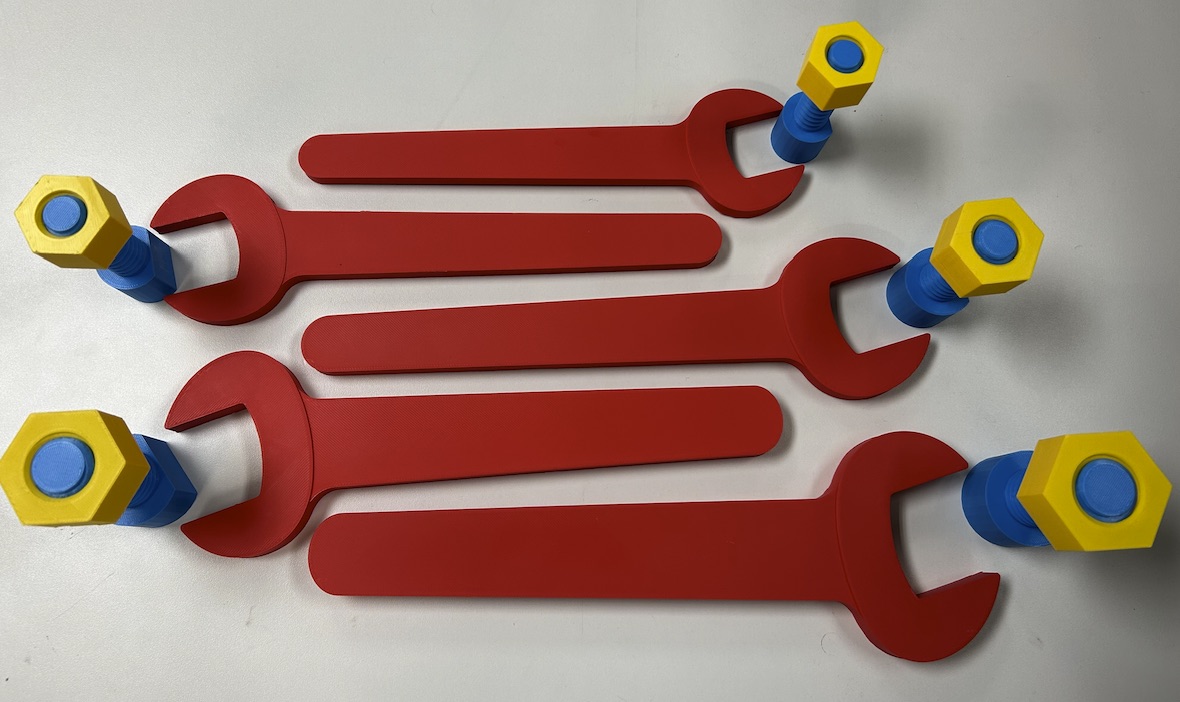}
    \caption{3D printed objects of the wrenches, nuts, and bolts. The tolerance between a wrench and its corresponding nut is 1-2 mm.}
    \label{fig:experiments-obj}
\end{figure}

The real-world setup, described in Sec.~\ref{method:sim2real}, is illustrated in Fig.~\ref{fig:real-setup}. The experiments are performed on five sets of wrench-nut-bolts of different sizes, shown in Fig.~\ref{fig:experiments-obj}. The side-to-side widths of the nuts range from 30 to 46 mm, and the openings of their corresponding wrenches are only 1-2 mm larger. We label these five assets by their increasing dimensions, from 1 to 5. Note that the heads of the wrenches used for sizes 4 and 5 are tilted by $19^\circ$ to the left, whereas those  used for sizes 1, 2 and 3 are tilted by $15^\circ$.

Each trial is initialized by randomly placing a bolt in a $10\times10$ cm square, $10$ cm above the table, with a random yaw orientation. The nut is inserted onto the bolt with a varying number of rotations. The Robotiq 3-finger gripper grasps the wrench's handle firmly and moves the wrench's head right above the nut. 

The insertion policy and the failure forecasting models are only trained with the size-5 assets in simulation. They are then tested on all five sets of assets (with four unseen sizes) in simulation and on three sets (sizes 1, 3, and 5) in the real-world experiments.

\begin{table*}[t]
    \centering    
    \caption{Insertion Success Rates, Total Number of Time Steps, and Reset Rates in The Real-World}
    \begin{tabular}{lccccccccc}
    \hline
    
    \multirow{2}{*}{Method} & \multicolumn{3}{c}{Size-5} & \multicolumn{3}{c}{Size-3} & \multicolumn{3}{c}{Size-1} \\
    \cline{2-4}\cline{5-7}\cline{8-10}
     & Succ. Rate & Steps & Reset Rate & Succ. Rate & Steps & Reset Rate & Succ. Rate & Steps & Reset Rate \\
     
    \hline
    IndustReal~\cite{tang2023industrealtransferringcontactrichassembly} & $30\%$ & $20$ & $-$
    & $37\%$ & $20$ & $-$
    & $23\%$ & $24$ & $-$\\

    Object-Centric & $94\%$ & $31$ & $-$
    & $92\%$ & $29$ & $-$
    & $92\%$ & $29$ & $-$\\

    \hline
    Time-Only & $100\%$ & $40$ & $14\%$
    & $96\%$ & $48$ & $27\%$
    & $94\%$ & $38$ & $13\%$\\

    Full-Trajectory & $\mathbf{100\%}$ & $34$ & $8\%$
    & $\mathbf{100\%}$ & $36$ & $14\%$
    & $\mathbf{96\%}$ & $31$ & $2\%$\\
    
    \hline
    \end{tabular} 
    \label{tab:real-single}
\end{table*}
\subsection{Evaluation in Simulation}

Table~\ref{tab:low-friction} presents the evaluation results of the IndustReal technique \cite{tang2023industrealtransferringcontactrichassembly} and the proposed object-centric method, with different failure prediction models, on the single-time wrench-nut insertion task for different asset sizes. The same randomization and observation noises described in  Sec.~\ref{method:sim2real} are used in all these tests. In this first set of experiments, we set the friction of the nut relative to the wrench to zero.

All the compared methods are tested with four random seeds, and with 128 trials for each seed. The mean and standard deviation of the insertion success rates are reported. The time limit $T$ of each insertion trial is $255$ time-steps. To examine the effect of the recovery behavior on the insertion time, we also list the average number of steps used in the successful trials.

The three proposed failure recovery methods (\textbf{Time-Only}, \textbf{Moving-Window}, and \textbf{Full-Trajectory}) are built on top of the proposed \textbf{Object-Centric} insertion policy. The number of actions executed by each recovery policy at each call is set to $T_R=30$. For the Moving-Window Success Model, the prediction window size and the cutoff threshold used during the evaluation are $T_F=30$ and $\alpha=0.13$, which are decided through a Bayesisan hyperparameter optimization~\cite{optuna_2019} after training the MLP.

Compared to IndustReal, the proposed object-centric insertion policy shows a $10\%$ increase in the success rate on Size-5 (seen during training), and generalizes consistently to all other unseen sizes while the former does not. The main insight here is that an object-centric pose representation eliminates task-agnostic noise, especially in tasks involving tool-object interactions. For the nut-insertion task considered in this paper, the poses of the bolt and the nut, as well as the grasping of the wrench, are irrelevant to the task completion. Therefore, their influences should be minimized in the observation space to reduce the exploration burden of the RL algorithm. 

As we can see in Table~\ref{tab:low-friction}, by adding the failure recovery module (Time-Only, Moving-Window, or Full-Trajectory) on top of our proposed object-centric policy, the insertion performance is further improved. 

\textbf{Model Robustness to Unexpected Physics} Simulating physical 
properties (e.g. the friction and deformation of objects) is challenging~\cite{narang2022factory}. However, in the real-world, these factors play an important role in our repeated high-precision manipulation task. To evaluate the robustness of the proposed method to unknown dynamics, we run experiments in the same simulated environment with the wrench-nut bolt friction set to {\bf 1}, where the nut has a higher tendency to rotate when touched by the wrench. The success rate and time steps are reported in Table~\ref{tab:high-friction}.

Overall, the system equipped with a recovery strategy is less affected by the high friction. By comparing the corresponding entries in Table~\ref{tab:high-friction} and Table~\ref{tab:low-friction}, we found that the drop of the success rate is around $3\%$ with recovery, and more than $5\%$ without recovery. This indicates that the simple retry strategy is effective and provides another layer of performance guarantee over the base policy. Among the three failure forecasting models, the learning-based success prediction models (Moving-Window and Full-Trajectory) consistently outperform the Time-Only model by spending an extra 11 time steps at most. While the Full-Trajectory Success model shows a slightly higher success rate than the Moving-Window Success model on 4/5 of the asset sets, the difference in performance is not significant. In practice, the Full-Trajectory Success model is preferred because the model input does not require the current time step $t$ (required by the Moving-Window Success Model), offering it the potential to adapt to a different control frequency in zero-shot without the need of aligning the observations with the time steps.

\subsection{Single-Time Insertion Experiments in Real-World}
Real-world experiments were conducted with the same embodiment as in the simulation: the Kuka iiwa14 7-DOF manipulator and the Robotiq 3-finger gripper. An Intel RealSense D435 RGB-D camera was utilized to track the object poses in the scene. The poses were used for autonomously grasping the wrench and initializing it above the nut with a programmed policy. The insertion policy and the failure recovery module then take over until the wrench head is inserted onto the nut, or reaching the time limit. The bolt was rigidly affixed to the table.

We evaluate the zero-shot sim-to-real transfer ability of the studied methods. Although IndustReal achieves significant performance in simulation, its robustness diminishes substantially in real-world deployment due to sim-to-real gaps. The proposed object-centric policy successfully transfers to the real robot. In the real experiments, we treat our object-centric policy as the non-recovery baseline, and compare it against two recovery methods Time-Only and Full-Trajectory. The Moving-Window does not transfer directly to the real robot because the control frequency is different. 

Note that the frequency of the position controller is roughly twice that frequency in simulation, so the number of time-steps in the real-world is half of that in simulation (with the same object-centric insertion policy). For this reason, the time limit of each trial $T$, as well as the consecutive steps $T_R$ of the recovery policy $\pi^R$, is shortened: $T=128, T_R=15$. 

Table~\ref{tab:real-single} reports three metrics of the single-time wrench-nut insertion experiments in real-world. Besides the success rate and the total time-steps, the reset rate refers to the occurrence of a recovery behavior in the successful trials. We intentionally chose to test on sizes 1, 3, and 5 because they are more challenging according to the evaluations in simulation. For each size and method, 50 real-world trials were collected, with randomly initialized nut and bolt poses (as explained in Sec.~\ref{exp:setup}).

The success rates of the recovery-aided methods are consistently higher than the non-recovery one. This confirms the necessity of failure forecasting in a robust system. The learned failure forecasting model (Full-Trajectory) takes fewer steps to success than the non-learning Time-Only model, because the neural network is able to forecast a failure earlier by observing the trajectory of the history. Further, the Full-Trajectory model presents a much lower reset rate while still maintaining a higher (or comparable) success rate, which is an indicator that the model correctly distinguishes between promising and struggling trials.

\subsection{Rhythmic Insertion Experiments in The Real-World}
These experiments quantitatively demonstrate the robustness of the nut-insertion system in real-world. Specifically, the robot is asked to continuously repeat 20 rounds of nut screwing. The system pipeline consists of four steps in each round:
\begin{enumerate}
    \item Initialization: placing the wrench head $\sim1cm$ above the nut and aligning the orientations;
    \item Insertion: inserting the wrench head onto the nut;
    \item Rotation: rotating the nut on the bolt counter-clockwise for at least $60^\circ$ using the wrench;
    \item Reset: lifting up the wrench and rotating back to the initial orientation.
\end{enumerate}
The insertion steps are performed by the insertion and recovery policies described above. The remaining three steps are accomplished by open-loop plans generated by predefined programs. 

\begin{figure}
    \centering
    \includegraphics[width=\linewidth]{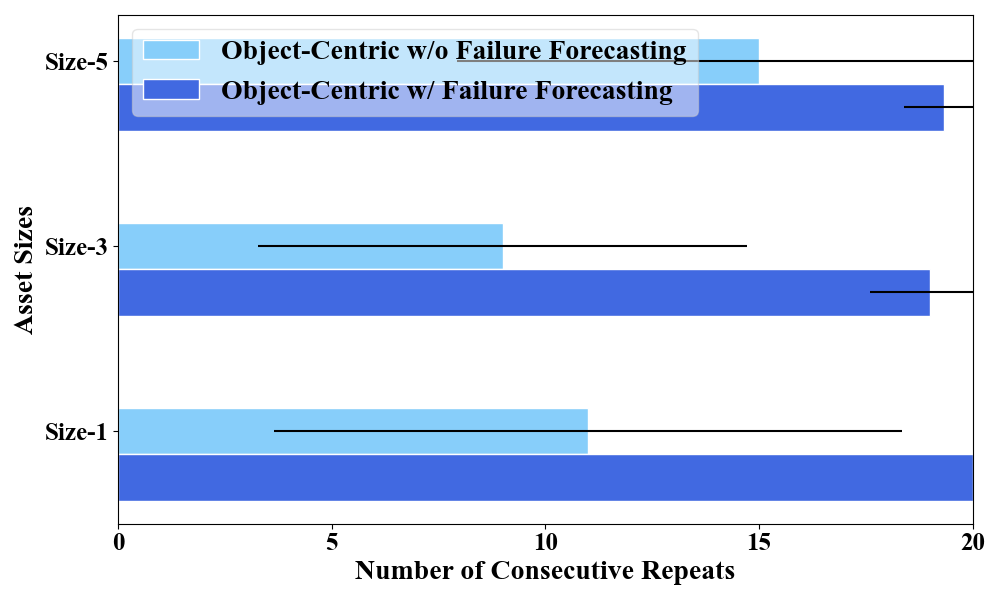}
    \caption{Number of consecutive insertion rounds over 3 real-world trials.}
    \label{fig:real-repeated}
\end{figure}

Fig.~\ref{fig:real-repeated} profiles the number of continuous insertion rounds with or without the recovery behavior. For each asset size, we conducted three trials with randomly initialized bolt and nut. The system is more persistent in this task if the bar ends closer to the right (i.e. 20 repeats). Clearly, our recovery-aided insertion policy contributes to a much more robust system in this rhythmic insertion task. 

\begin{figure*}
    \centering
    \includegraphics[width=1\linewidth]{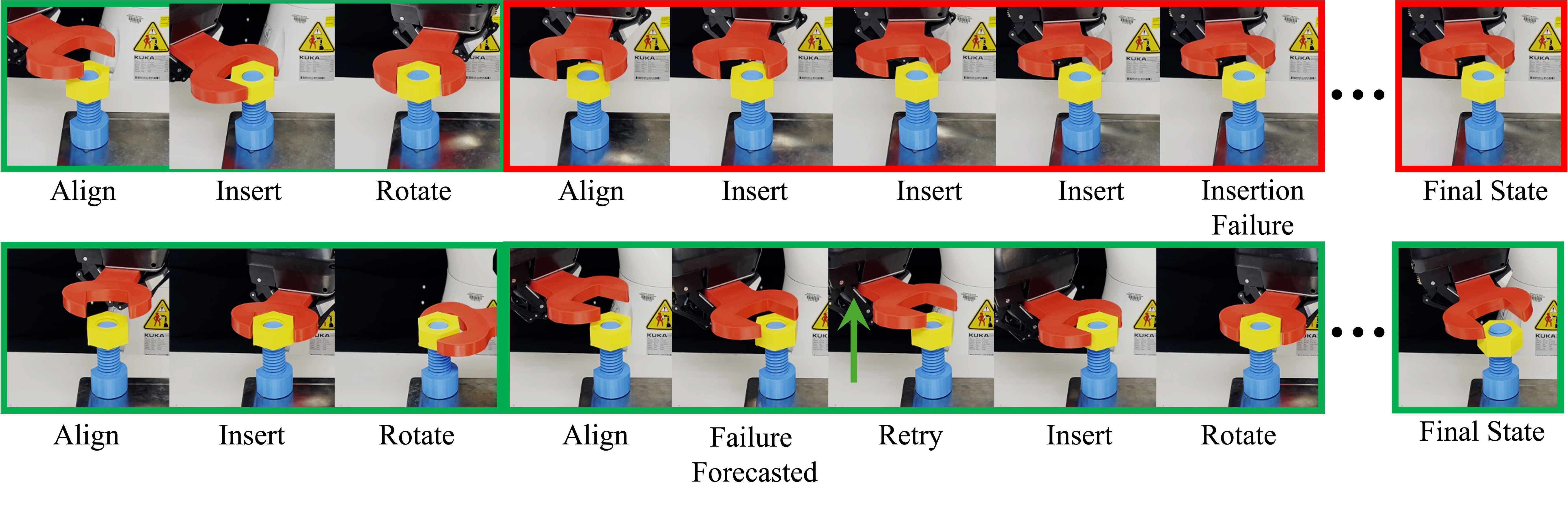}
    \vspace{-0.9cm}
    \caption{Snapshots of two consecutive rounds from the rhythmic insertion experiments on the size 5 assets. \textbf{Top Row}: The non-recovery insertion policy. \textbf{Bottom Row}: The same insertion policy with recovery guided by the learned failure forecasting model.}
    \label{fig:real-repeated-vis}
    \vspace{-0.5cm}
\end{figure*}

This performance boost can be justified with a simple statistical calculation. Let $p$ denote the success probability of a single-time insertion and assume the repeated insertions are independent in this task. We want to find the number of consecutive successes $N$ before the first failure. This random variable $N$ can be modeled by the geometric distribution, and the expected value of $N$ is calculated by:
\begin{equation}
    \mathbf{E}(N) = \frac{p}{1-p}.
\end{equation}
We can estimate the success probability $p$ with the empirical results of Table~\ref{tab:real-single}. Take size 1 assets as an example. The single-time insertion success probabilities and the expected number of consecutive successes are listed in Table \ref{tab:real-repeated-analysis}.
\begin{table}[h]
    \centering
    \caption{Statistical Analysis of Rhythmic Insertion Experiments in Real-World}
    \begin{tabular}{|c|c|c|}
    \hline
        Method & $p$& $\mathbf{E}(N)$\\
        \hline
         Object-Centric (non-recovery) & $92\%$& 11.5\\
         Full-Trajectory (recovery) & $96\%$& 24\\
         \hline
    \end{tabular}
    \label{tab:real-repeated-analysis}
\end{table}

The expected consecutive successes match nicely with our empirical results in Fig.~\ref{fig:real-repeated}.
Qualitative results of this repeated insertion task are shown in Fig.~\ref{fig:real-repeated-vis}, where representative snapshots of two consecutive nut insertion rounds are visualized for both methods. The non-recovery method encountered an insertion failure at an early round, leading to a barely fastened nut. On the hand, the same insertion policy equipped with a recovery mechanism managed to recover early from such insertion failures and it succeeded after retrying. After 20 rounds of rotation, the nut went down by a centimeter. The process continues until the nut is fully inserted.

\section{Conclusion}
We present a robust policy specifically designed for {\it Rhythmic Insertion Tasks}, which requires uninterrupted succession of insertions with high precision. Our method consists of two components: a single-insertion reinforcement learning policy operating on the object-centric pose representations, and a failure recovery module that generates retry actions when a future failure is predicted by our learned failure forecasting model. Both the insertion policy and the failure forecasting model are trained exclusively in simulation and transferred in zero-shot to the real world. Extensive experiments in simulation and the real world clearly demonstrate the robustness of the proposed method for the single-insertion task, and the tests on nut-insertion with a wrench demonstrate the necessity of failure prediction and recovery in RIT.

\bibliographystyle{IEEEtran}
\bibliography{sections/reference} 

\addtolength{\textheight}{-12cm}   


\end{document}